\documentclass[10pt,twocolumn,letterpaper]{article}

\usepackage{cvpr}              

\usepackage{booktabs}
\usepackage{multirow}
\usepackage{caption}
\usepackage{graphicx}
\usepackage{subcaption}
\usepackage[accsupp]{axessibility} 

\usepackage{CJKutf8}
\usepackage{graphicx}

\definecolor{cvprblue}{rgb}{0.21,0.49,0.74}
\usepackage[pagebackref,breaklinks,colorlinks,allcolors=cvprblue]{hyperref}

\def\paperID{36816} 
\def\confName{CVPR}
\def\confYear{2026}
\title{CRFT: Consistent–Recurrent Feature Flow Transformer \\ for Cross-Modal Image Registration}

\author{
Xuecong Liu\textsuperscript{1} \quad
Mengzhu Ding\textsuperscript{1} \quad
Zixuan Sun\textsuperscript{2} \quad
Zhang Li\textsuperscript{2} \quad
Xichao Teng\textsuperscript{2}\thanks{Corresponding author.} \\
\textsuperscript{1}Northeastern University, China \quad
\textsuperscript{2}National University of Defense Technology, China \\
{\tt\small liuxuecong@qhd.neu.edu.cn} \quad {\tt\small tengari@buaa.edu.cn}
}

\begin{document}
\maketitle
\begin{abstract}
\begin{CJK*}{UTF8}{gbsn}

We present Consistent–Recurrent Feature Flow Transformer (CRFT), a unified coarse-to-fine framework based on feature flow learning for robust cross-modal image registration. CRFT learns a modality-independent feature flow representation within a transformer-based architecture that jointly performs feature alignment and flow estimation. 
The coarse stage establishes global correspondences through multi-scale feature correlation, while the fine stage refines local details via hierarchical feature fusion and adaptive spatial reasoning. To enhance geometric adaptability, an iterative discrepancy-guided attention mechanism with a Spatial Geometric Transform (SGT) recurrently refines the flow field, progressively capturing subtle spatial inconsistencies and enforcing feature-level consistency. This design enables accurate alignment under large affine and scale variations while maintaining structural coherence across modalities. Extensive experiments on diverse cross-modal datasets demonstrate that CRFT consistently outperforms state-of-the-art registration methods in both accuracy and robustness. Beyond registration, CRFT provides a generalizable paradigm for multimodal spatial correspondence, offering broad applicability to remote sensing, autonomous navigation, and medical imaging. 
Code and datasets are publicly available at 
\url{https://github.com/NEU-Liuxuecong/CRFT}.

\end{CJK*}
\end{abstract}    
\section{Introduction}
\label{sec:intro}

Establishing reliable spatial correspondence between images captured under heterogeneous sensing conditions is a central problem in computer vision~\cite{jiang2021review,yang2023towards,gehrig2024low,chen2025survey,li2025object}. 
It supports a wide range of applications, including 3D reconstruction~\cite{edstedt2025colabsfm, leroy2024grounding}, visual localization~\cite{liu2022fast}, autonomous navigation~\cite{wang2024survey}, and remote sensing analytics~\cite{Yu_2025_ICCV, ye20253mos}. 
In practice, images may originate from different modalities (e.g., optical, infrared, SAR, multispectral~\cite{li2024sardet, Qin_2025_CVPR, Poggi_2025_ICCV}) or be acquired under distinct viewpoints, focal lengths, or illumination settings~\cite{zhang2025adapting, schusterbauer2025diff2flow}, resulting in substantial appearance and geometric variability.

Cross-modal and deformable registration~\cite{mao2025cross,li2025implicit} is particularly difficult because the underlying sensing mechanisms introduce severe nonlinear appearance discrepancies and complex geometric variations~\cite{leroy2024grounding, chen2024mvsplat, wang2024recursive, wu2024single, zhang2025comatcher}. Differences in imaging intensity and phase contrast between optical and SAR sensors, as well as the spectral gap between visible and infrared bands, lead to inconsistent feature distributions. Meanwhile, geometric transformations \cite{Dong_2025_CVPR, Zhao_2025_CVPR} caused by camera motion, scale change, or viewpoint variation further complicate pixel-level alignment~\cite{bie2025graphi2p}, rendering traditional feature-based or intensity-driven methods unreliable \cite{archana2024deep,wu2025dfm, dai2024dsap}. 

To address these challenges, we propose CRFT, a unified coarse-to-fine framework designed to learn modality-independent feature flow for robust cross-modal image registration. At the coarse level, CRFT constructs a global correspondence field through multi-scale feature correlation, enabling reliable alignment even when only weak structural cues are available. At the fine level, we introduce a hierarchical feature fusion process that enhances local details and texture precision. Furthermore, to handle spatial distortions and geometric deformations, we design an iterative discrepancy-guided attention mechanism coupled with a SGT module. This recurrent refinement strategy progressively updates the flow field by dynamically weighting feature discrepancies, achieving pixel-wise alignment with high geometric fidelity.

Our contributions are summarized as follows:

\textbf{(1)} We introduce CRFT, a unified transformer-based framework that formulates cross-modal image registration as a learnable feature flow estimation task, jointly learning modality-independent representations and dense spatial correspondences.

\textbf{(2)} A coarse-to-fine hierarchical matching strategy is developed, where the coarse stage captures global context using low-resolution correlation, while the fine stage performs local refinement with high-resolution feature fusion, achieving precise hierarchical alignment.

\textbf{(3)} An iterative discrepancy-guided attention mechanism integrates feature discrepancy weighting with SGT and flow-based recurrent updates, progressively refining spatial alignment and enhancing robustness to nonlinear and affine geometric variations.
\section{Related Works}
\label{sec:relatedworks}

\noindent \textbf{Hand-Crafted Matching.}
Classical multimodal registration relies on region-based or feature-based handcrafted descriptors. 
Region-based methods~\cite{xuecong2024robust, liu2024shape, teng2023omird} achieve high precision but are sensitive to rotation, scale, and affine variations. 
Feature-based approaches such as RIFT~\cite{li2019rift}, RIFT2~\cite{li2023rift2}, LNIFT~\cite{li2022lnift}, HOWP~\cite{zhang2023histogram}, MSG~\cite{zheng2025msg}, and OS-SIFT~\cite{xiang2018ossift} design modality-invariant descriptors, yet struggle under strong nonlinear intensity shifts or low-texture conditions.

\noindent \textbf{Sparse Learnable Matching.}
Learned detectors and descriptors (e.g., SuperPoint~\cite{detone2018superpoint}, SuperGlue~\cite{sarlin2020superglue}, LightGlue~\cite{lindenberger2023lightglue}, OmniGlue~\cite{jiang2024omniglue}) improve sparse matching by integrating contextual reasoning via CNNs and transformers.  
While effective for RGB scenarios, their reliance on keypoint-descriptor coupling limits generalization to cross-modal settings where positional statistics differ significantly.

\noindent \textbf{(Semi-)Dense Learnable Matching.}
Detector-free or semi-dense models~\cite{xiao2024adrnet, lee2025dense, xu2023murf} learn denser correspondences. 
LoFTR~\cite{sun2021loftr} and its variants (XoFTR~\cite{tuzcuouglu2024xoftr}, ELoFTR~\cite{wang2024efficient}) combine coarse global attention with fine refinement, improving robustness to viewpoint change. 
ReDFeat~\cite{deng2022redfeat} extends dense descriptors to cross-domain data.  
However, many of these methods are still optimized on homogeneous RGB datasets, and their performance degrades when facing large geometric distortions or severe cross-modal appearance gaps \cite{potje2024xfeat}.

\noindent \textbf{Flow-Based Matching.}
Optical-flow-inspired networks \cite{xing2025goalflow, Morimitsu_2025_CVPR, Li_2025_ICCV} have been widely adopted for dense correspondence estimation. RAFT introduced an effective recurrent update paradigm that iteratively refines a flow field via a correlation volume and update operator, setting a new standard for dense flow. Subsequent works \cite{xu2022gmflow, xu2023unifying, huang2022flowformer, shi2023flowformer++, edstedt2024roma, xu2023murf} combine transformer-style global reasoning with recurrent refinement or incorporate memory/attention modules to handle large displacements and complex motion~\cite{liu2026gaff, sun20243}. However, most flow models assume photometric consistency and are trained on RGB optical-flow benchmarks; thus, they do not directly account for cross-modal appearance mismatches without modification to handle modality gaps or to learn modality-robust feature spaces.

\noindent \textbf{Geometry-Aware and Refinement.}
A growing line of work integrates explicit geometric modeling or spatial transforms \cite{xiao2024adrnet, xu2023murf, edstedt2023dkm, edstedt2024roma} (e.g., spatial transformer networks, learned warping layers) with feature learning to improve robustness to viewpoint, scale, and affine variations.
Several advanced flow-based methods \cite{fu2025moflow} employed distinct mechanisms for iterative refinement. 
RAFT \cite{teed2020raft}
progressively updates flow fields through sequential GRU steps, enabling high-precision estimation even in scenarios with large displacements. 
GMFlow \cite{xu2022gmflow}
integrates attention mechanisms into its optimization framework. 
FlowFormer \cite{huang2022flowformer} 
proposes a recurrent cost decoder that iteratively refines estimated optical flows. 
GDROS \cite{sun2025gdros} utilizes a geometry-guided mechanism for cross-modal optical flow, based on Least Squares Regression.

CRFT leverages global context to establish robust coarse correspondences, it performs dense, recurrent updates for sub-pixel accuracy. Crucially, CRFT departs from traditional optical-flow assumptions by explicitly learning a modality-independent feature flow within a transformer backbone, and by introducing a discrepancy-guided attention mechanism coupled with a SGT to iteratively correct geometric misalignments. This approach bridges the gap between global geometric consistency and local precision, offering a scalable solution for registration tasks.

\section{Methodology}
\label{sec:Methodology}

\begin{figure*}[t]
    \centering
    \includegraphics[width=1\textwidth]{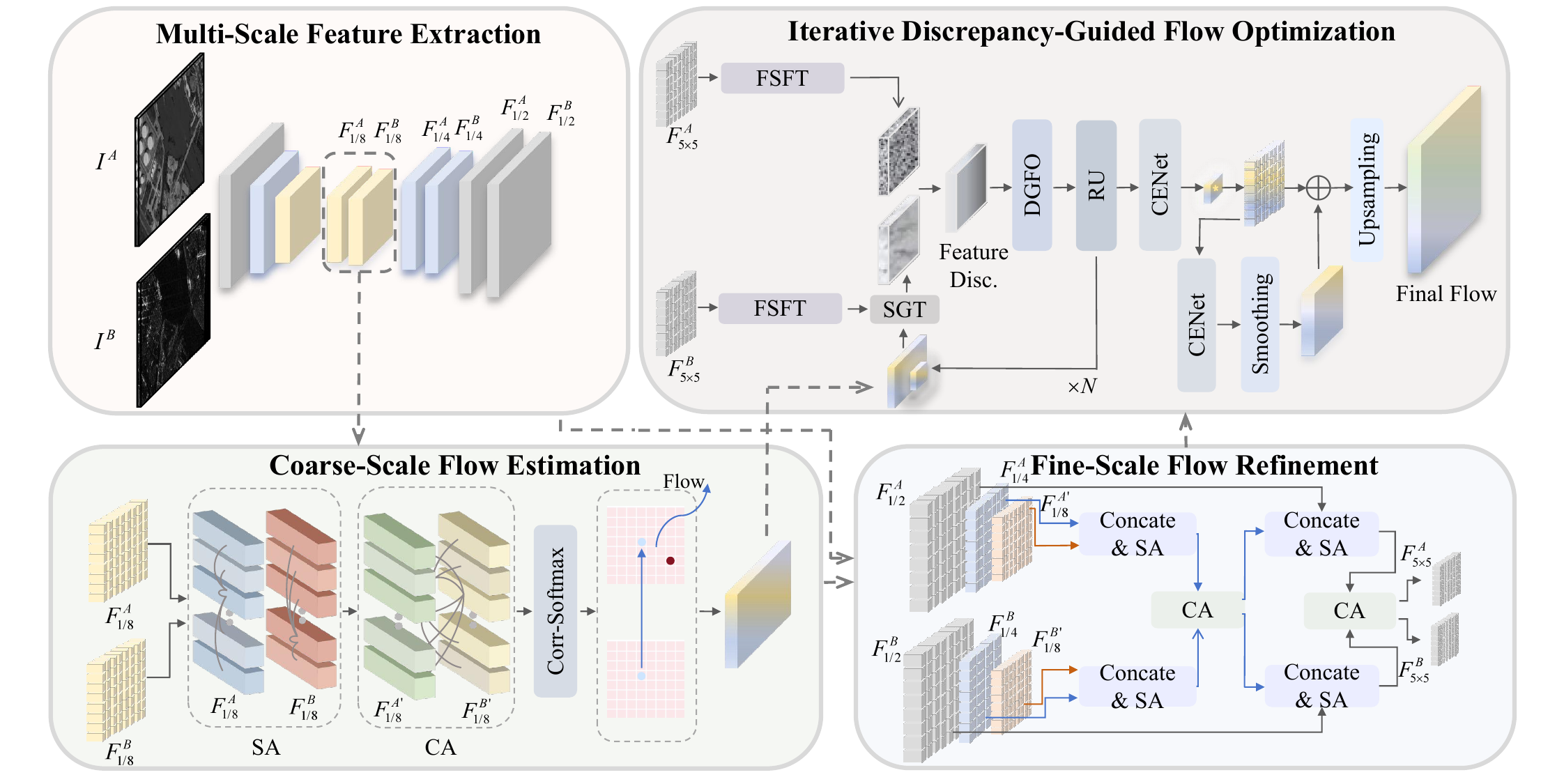} 
    \caption{
    \textbf{Overview of the proposed CRFT framework.} 
    CRFT follows a unified coarse-to-fine pipeline for robust cross-modal image registration. 
    (a) \textbf{Multi-Scale Feature Extraction:} A CNN encoder extracts modality-independent features at $\{1/2,1/4,1/8\}$ resolutions from the input image pair $(I^{A}, I^{B})$. 
    (b) \textbf{Coarse-Scale Flow Estimation:} At 1/8 resolution, transformer-style self-attention (SA) and cross-attention (CA) operate on low-resolution features to compute global correlation and generate an initial coarse flow field. 
    (c) \textbf{Fine-Scale Flow Refinement:} The 1/2 and 1/4 features are fused through window-based self-attention and cross-attention to inject fine-grained spatial details and locally refine coarse correspondences. 
    (d) \textbf{Iterative Discrepancy-Guided Flow Optimization:} Within $N$ refinement iterations, fine-scale features undergo Fine-Scale Feature Transformation (FSFT), aligned via a Spatial Geometric Transform (SGT), and compared to compute feature discrepancies. These discrepancies guide the Discrepancy-Guided Flow Optimization (DGFO), followed by a Residual Update (RU) and a Confidence Estimation Network (CENet) for smoothing, producing the final high-precision dense flow.}    
    \label{fig:overview}
\end{figure*}

CRFT is a unified coarse-to-fine framework designed to learn
modality-independent feature flow for robust cross-modal image registration. 
Unlike optical-flow architectures such as RAFT or GMFlow that rely on photometric consistency and often collapse under modality gaps, CRFT explicitly models multi-scale structure, cross-modal feature interactions, and geometric constraints. 
An overview of our framework is shown in Fig.~\ref{fig:overview}.

\subsection{Coarse-Scale Flow Estimation}
The first stage, flow-based coarse matching, aims to generate the initial flow field at coarse scales. Operating at a larger spatial scale, this stage facilitates the effective capture of dominant texture structures within the images. The process involves multi-scale feature extraction at 1/2, 1/4, and 1/8 resolutions, feature enhancement through a Transformer-based module, and initial flow estimation, where global correlation between the coarse-scale features $F^{A}_{1/8} $and $F^{B}_{1/8}$ is computed to predict coarse correspondence coordinates and obtain the initial optical flow.

\noindent\textbf{Multi-scale Feature Extraction.}
A ResNet-based CNN encoder shared across modalities extracts features at 1/2, 1/4, and 1/8 scales.  
Coarse matching is performed at the 1/8 scale, where features capture high-level structures less affected by spectral or radiometric inconsistency, while 1/2 and 1/4 features retain finer spatial details for later refinement.  
This multi-scale separation is essential: coarse structural cues enable stable global matching despite modality gaps, while higher-resolution features avoid the drift typically observed in coarse-only matching.

\noindent\textbf{Coarse-Scale Feature Transformation:}
To mitigate the inherent cross-modal discrepancy, we introduce transformer blocks consisting of self-attention (SA) and cross-attention (CA).  
Positional encodings are first added to the 1/8-scale features $F^{A}_{1/8}$ and $F^{B}_{1/8}$, after which repeated SA-CA layers aggregate long-range context within each modality and fuse complementary 
structural cues across modalities.  
This design is particularly effective for cross-modal matching: SA stabilizes pixel-to-context embedding, while CA explicitly aligns modality-dependent features, enabling transformation-invariant global descriptors that outperform purely correlation-based methods.

\noindent
\textbf{Flow Initialization via Global Correlation.}
Subsequent to feature transformation, we apply linear projection and normalization to $F^{A}_{1/8}$ and $F^{B}_{1/8}$. For each feature vector $f\in R^{C}$, the transformed feature $\hat{f}$ is computed as follows:

\begin{equation}
  \hat{f} = \frac{W_{proj}\cdot f+b_{proj}}{\sqrt{C}},
  \label{eq:important}
\end{equation}

\noindent
where $W_{proj}$ and $b_{proj}$ are the learnable parameters of the linear projection layer, and \textit{C} is the feature dimension.

Global correlation is calculated to connect the feature from $\hat{F^A}$, $\hat{F^B}\in R$, the correlation is calculated as follows:

\begin{equation}
    \mathcal{Corr}(p,q)=\left\langle\hat{F^A}(p),\hat{F^B}(q)\right\rangle,
\end{equation}
where $(p,q)$ denotes pixel locations. 
Matching probabilities are obtained via softmax normalization:
\begin{equation}
 P(p,q)=\frac{\exp(\mathcal{Corr}(p,q))}{\sum_{q'}\exp(\mathcal{Corr}(p,q'))}.
\end{equation}

The coarse correspondence and the coarse flow can be computed as:
\begin{equation}
    p'=\sum_{q}P(p,q)\cdot q,
\end{equation}

\begin{equation}
    T_c(p)=p'-p.
\end{equation}

This global matching strategy provides a geometry-aware, modality-insensitive initial flow that remains stable even under large radiometric differences, where optical-flow baselines typically fail.

\subsection{Fine-Scale Flow Refinement}

The coarse flow provides globally reliable but spatially imprecise correspondences due to the low resolution and strong modality discrepancy. 
To recover fine-grained alignment, CRFT introduces a hierarchical fine-scale refinement module that progressively injects high-resolution structural cues while maintaining cross-modal consistency.

\noindent\textbf{Multi-scale Local Feature Gathering.}
For each location in the coarse feature map, we extract local feature windows from the 1/4- and 1/2-resolution feature maps using $5\times5$ and $3\times3$ kernels, respectively. 
This establishes explicit cross-scale correspondence paths: each coarse-level pixel is associated with mid- and fine-scale neighborhoods containing richer spatial and textural information.  
Such hierarchical windowing is crucial for cross-modal matching, as it provides modality-invariant structural context at multiple resolutions without relying on local intensity patterns.

\noindent\textbf{Hierarchical Attention-based Refinement.}
Coarse- and mid-scale windows are first refined using windowed self-attention to enhance local geometric consistency. 
Cross-attention is then applied to the $3\times3$ mid-scale windows to align modality-specific responses at finer scales.  
These refined features are fused with the $1/2$-resolution windows through the same SA-CA operations, allowing progressively finer details to be injected across scales.  
This hierarchical process produces a multi-scale fused representation that preserves coarse structural reliability while supplying the high-frequency cues needed for precise sub-pixel flow estimation, effectively bridging global coarse matching and dense fine-level alignment.

\subsection{Discrepancy-Guided Flow Optimization}

\begin{figure}[t]
    \centering
    \includegraphics[width=0.5\textwidth]{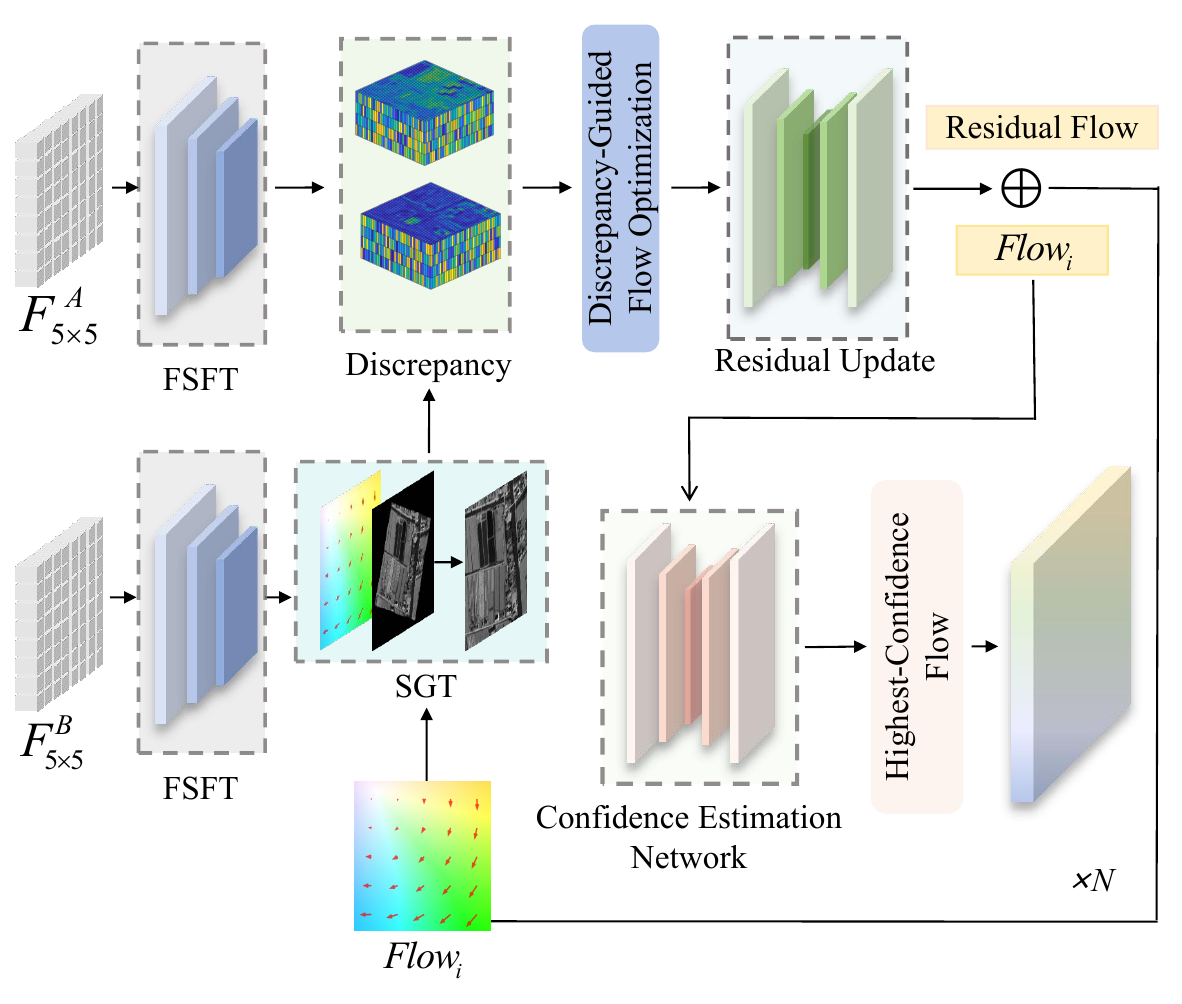} 
    \caption{\textbf{Iterative discrepancy-guided flow refinement.}
    At each iteration, fine-scale features from both modalities are first mapped into a shared feature space (FSFT).  
    The current flow then drives a SGT to align the local window of $F_{5\times5}^B$ with $F_{5\times5}^A$.  
    Feature discrepancies between the aligned windows are computed and used to weight an attention-based refinement module, which predicts a residual flow added to the current estimate.  
    A confidence network further smooths the updated flow before the next iteration.  
    This geometry-aware, discrepancy-driven process progressively improves cross-modal alignment.}
    \label{fig:iterativve}
\end{figure}

While the coarse and fine matching modules provide an initial feature flow, accurately aligning cross-modal images requires a mechanism that can continuously correct local geometric errors and modality-induced inconsistencies. 
To this end, we introduce an iterative discrepancy-guided optimization module coupled with a Spatial Geometric Transform (SGT).  
This module constitutes the core innovation: it performs recurrent flow refinement inside local windows, explicitly leveraging feature discrepancies to steer geometry-aware flow updates—an ability that conventional optical-flow networks lack under modality gaps.

\noindent\textbf{Fine-Scale Feature Transformation (FSFT).}
Before the iterative refinement, the fine-level features from the previous stage are projected into a homogeneous feature space using a lightweight MLP consisting of two linear layers and a GELU activation.  
This step significantly reduces the appearance discrepancy between modalities and stabilizes the subsequent discrepancy estimation.

\noindent\textbf{Discrepancy-Guided Flow Optimization (DGFO).}
Figure~\ref{fig:iterativve} illustrates the proposed iterative refinement strategy.  
Given the initial coarse flow $T_c$, we expand the $1/8$-scale flow to match the spatial support of the $5\times5$ window at $1/2$ resolution, forming the initialization $T_{k_c}^0$:
\begin{equation}
    T_k^i(p)=T_{k_c}^{i-1}, \quad p\in[0,W_f-1]^2.
\end{equation}

For each iteration, a coordinate grid is generated for $I^B$, and the current flow $T_k^i$ is used to warp the features via SGT, aligning the local neighborhood in $I^A$ with its counterpart in $I^B$.  
This yields aligned feature pairs whose discrepancy is defined as:
$ \Delta F_{k}^{i} = \text{Normalize}\!\left(F_A - \mathcal{W}(F_B, T_{k}^{i})\right), $
where $\mathcal{W}$ is the spatial geometric transform (SGT) that warps target features using the current flow estimate.
The inverted discrepancy is:
\begin{equation}
    F_{attn}=1-\Delta F_k^i,
\end{equation}
This inverted discrepancy serves as a reliability cue—larger consistency leads to stronger attention responses. The discrepancy-modulated feature map produces the queries and keys:
\begin{equation}
    Q(p)=W_QF_{attn}(p), \qquad K(q)=W_KF_{attn}(q),
\end{equation}
while the current flow $T_k^i$ is treated as the value.  
Local attention is computed within the neighborhood $\mathcal{N}(p)$:
\begin{equation}
    \alpha_{p,q}=\frac{\exp(\langle Q(p),K(q)\rangle/\sqrt{C_f})}{\sum_{q'\in \mathcal{N}(p)}\exp(\langle Q(p),K(q')\rangle/\sqrt{C_f})},
\end{equation}
where $C_f$ is the feature channel dimension of $Q$ and $K$. The aggregated flow is:

\begin{equation}
    T_k'^i(p)=\sum_{q\in \mathcal{N}(p)}\alpha_{p,q}\,T_k^i(q).
\end{equation}

\noindent\textbf{Residual Update (RU).}
The aggregated flow is passed through a CNN-Encoder to estimate a residual update. By adding the residual flow$\Delta T_k^i$ to $T_k'^i$, the $T_k^{i+1}$ is returned as the update flow for this iteration, yielding:
\begin{equation}
    T_k^{i+1}=T_k'^i + \Delta T_k^i.
\end{equation}

\noindent\textbf{Local Flow Aggregation.}
Each iteration employs a confidence estimation network (CENet), which is predicted from both the refined features and the original image content, to compute a confidence map for the current local flow window. Within each local window, the highest-confidence flow is selected and aggregated to form a coherent flow field.

\noindent\textbf{Confidence Smoothing.}
The refined flow field undergoes confidence smoothing, during which the recomputed confidence map guides a weighted smoothing:
\begin{equation}
    T'_f = \alpha\cdot \textit{ConvNet}_{smooth}(T_f \odot M_f) + (1-\alpha)\cdot T_f.
\end{equation}
Finally, this flow field is upsampled to the full image resolution, generating a high-precision flow estimation.

Through repeated refinement iterations, CRFT progressively corrects geometric distortions and cross-modal inconsistencies, ultimately producing a dense, high-fidelity registration field. 
Unlike conventional optical-flow refinement, our discrepancy-guided, geometry-aware optimization adapts to modality gaps and achieves stable convergence even under large spectral or radiometric variations.

\subsection{Loss Function}

CRFT is trained in a multi-stage fashion, and we design a corresponding hierarchical loss to supervise both coarse-level correspondence estimation and iterative fine-level refinement. The loss consists of two components: a coarse-scale flow loss and a geometrically weighted iterative refinement loss.

\noindent\textbf{Coarse-Level Flow Supervision.}
To guide the model toward learning globally consistent correspondences, the initial coarse flow $T_c$ is supervised by a downsampled ground-truth flow $T_{gt,c}$.  
This encourages the network to capture large-scale structural alignment that provides a robust initialization for the subsequent refinement stages:
\begin{equation}
\mathcal{L}_c = \frac{1}{|\mathcal{V}_c|} 
\sum_{p \in \mathcal{V}_c} \left\| T_c(p)-T_{gt,c}(p) \right\|_1 .
\end{equation}

\noindent\textbf{Iterative Fine-Level Supervision.}
During iterative refinement, a flow prediction $T_f^i$ is produced at each iteration.  
Each iteration is supervised independently using an $L_1$ flow loss:
\begin{equation}
\mathcal{L}_f^i = \frac{1}{|\mathcal{V}|} 
\sum_{p \in \mathcal{V}} \left\| T_f^i(p)-T_{gt}(p) \right\|_1.
\end{equation}

To reflect the progressive nature of the refinement and encourage convergence toward high-precision alignment, we aggregate all iteration losses using geometrically decaying weights:
\begin{equation}
\mathcal{L}_f = \sum_{i=1}^{N} \gamma^{N-i}\, \mathcal{L}_f^i ,
\end{equation}
where $\gamma=0.9$ controls the decay rate, assigning larger weights to later (more accurate) predictions.

\noindent\textbf{Total Loss.}
The total loss is a weighted combination of the coarse loss and the iterative refinement loss:
\begin{equation}
\mathcal{L}_{total} 
= \lambda_c\, \mathcal{L}_c 
+ \lambda_f\, \mathcal{L}_f.
\end{equation}

This hierarchical weighting scheme provides strong gradient signals for early iterations while ensuring that the most accurate late-stage predictions dominate supervision.  
As a result, the refinement process is explicitly driven toward high-precision convergence—an ability that is critical for cross-modal registration and is not provided by single-stage or uniformly weighted loss formulations.

\section{Experiments}
\label{sec:experiments}



\subsection{Implementation Details}

We initialize CRFT using the publicly released XoFTR pretrained weights (trained at 640 resolution) and fine-tune the entire framework end-to-end. 
Training is performed on OSdataset and RoadScene, a batch size of 16, and the AdamW optimizer with a learning rate of $3\times 10^{-4}$.




\noindent\textbf{Datasets.}
We evaluate CRFT on two representative cross-modal benchmarks: OSdataset~\cite{xiang2020automatic} with 2673 aligned optical–SAR pairs, and RoadScene~\cite{xu2020u2fusion} with 221 aligned visible–infrared road-scene pairs. We follow the standard splits of 2011/238/424 (train/val/test) for OSdataset and 176/23/22 for RoadScene. All images are resized to $512\times512$ before patch sampling. 
To improve robustness to geometric variations, training images are augmented with random scaling in $[0.9,1.1]$, rotations in $[-45^\circ,45^\circ]$, and up to 30-pixel translations. During training, we randomly crop $64\times64$ patches, and ground-truth flow fields are generated directly from the provided pixel-level alignments for consistent supervision across all stages.

\noindent\textbf{Evaluation Metrics.}
We evaluate registration accuracy using two standard metrics: 
Average End-Point Error (AEPE), which measures the mean flow prediction error, 
and Correct Match Rate (CMR), which reports the percentage of samples whose AEPE falls below a threshold~$\tau$. 
We follow prior works by computing CMR across multiple thresholds ($0.1\!-\!5$ px), providing a comprehensive view of performance under different geometric tolerances.

\noindent\textbf{Baselines.}
Following ELoFTR~\cite{wang2024efficient} and MINIMA~\cite{ren2025minima}, we compare CRFT with representative methods from three major categories:

\begin{itemize}
    \item \textit{Sparse matching methods:}  
    HOWP~\cite{zhang2023histogram},  
    LNIFT~\cite{li2022lnift},  
    MSG~\cite{zheng2025msg},  
    RIFT2~\cite{li2019rift, li2023rift2}.     
    These handcrafted keypoint-descriptor pipelines are robust to structural distortions but fail under severe modality gaps or large geometric variations.

    \item \textit{Semi-dense matching methods:}  
    XoFTR~\cite{tuzcuouglu2024xoftr}.  
    It represents modern transformer-based detector-free matchers, but its performance degrades when cross-modal texture distributions diverge.

    \item \textit{Dense matching methods:}  
    ADRNet~\cite{xiao2024adrnet},  
    GMFlow~\cite{xu2023unifying},  
    RAFT~\cite{teed2020raft},  
    GDROS~\cite{sun2025gdros}.     
    These approaches directly estimate pixelwise correspondences, but traditional optical-flow models fail under strong modality discrepancy, while GDROS and ADRNet rely on modality-specific priors and limited resolution.
\end{itemize}

Together, these baselines cover handcrafted, transformer-based, and dense optical-flow paradigms, providing a comprehensive and challenging benchmark to demonstrate the effectiveness of CRFT.

\subsection{Experiment Results}

\begin{figure}[t]
    \centering
    \includegraphics[width=0.5\textwidth]{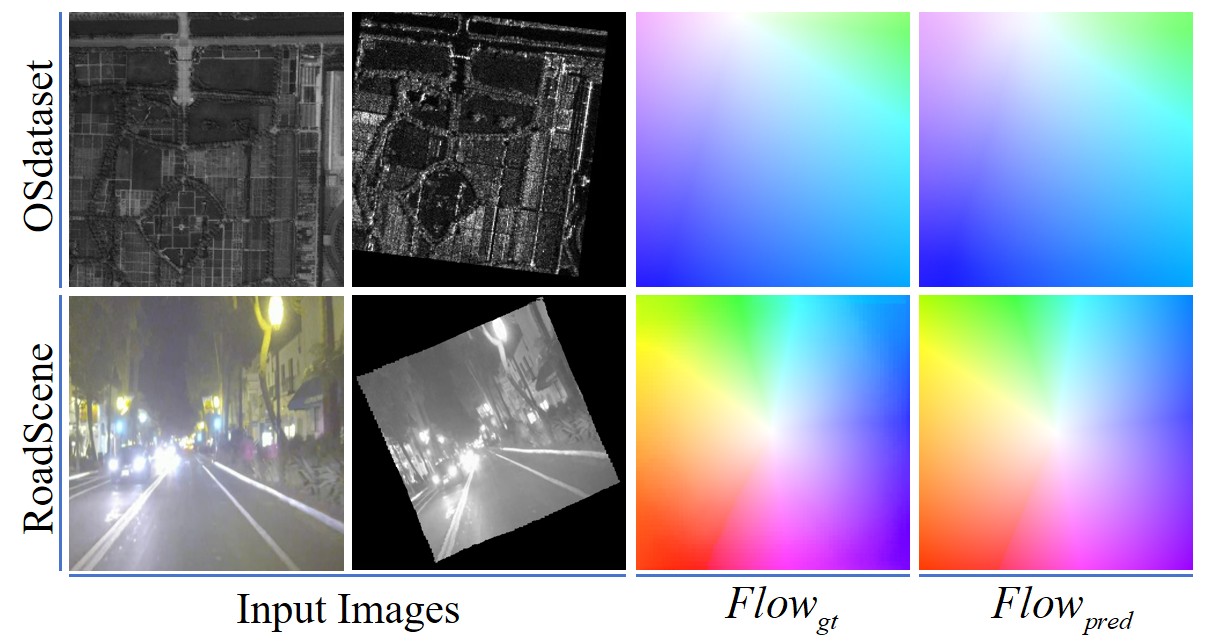} 
    \caption{\textbf{Visual comparison between the predicted flow and ground-truth flow fields.} The proposed CRFT achieves geometrically consistent and dense alignment across challenging optical-SAR and optical-infrared image pairs, demonstrating strong robustness to nonlinear radiation and geometric variations.}
    \label{fig:flow}
\end{figure}

\begin{figure*}[t]
    \centering
    \includegraphics[width=0.9\textwidth]{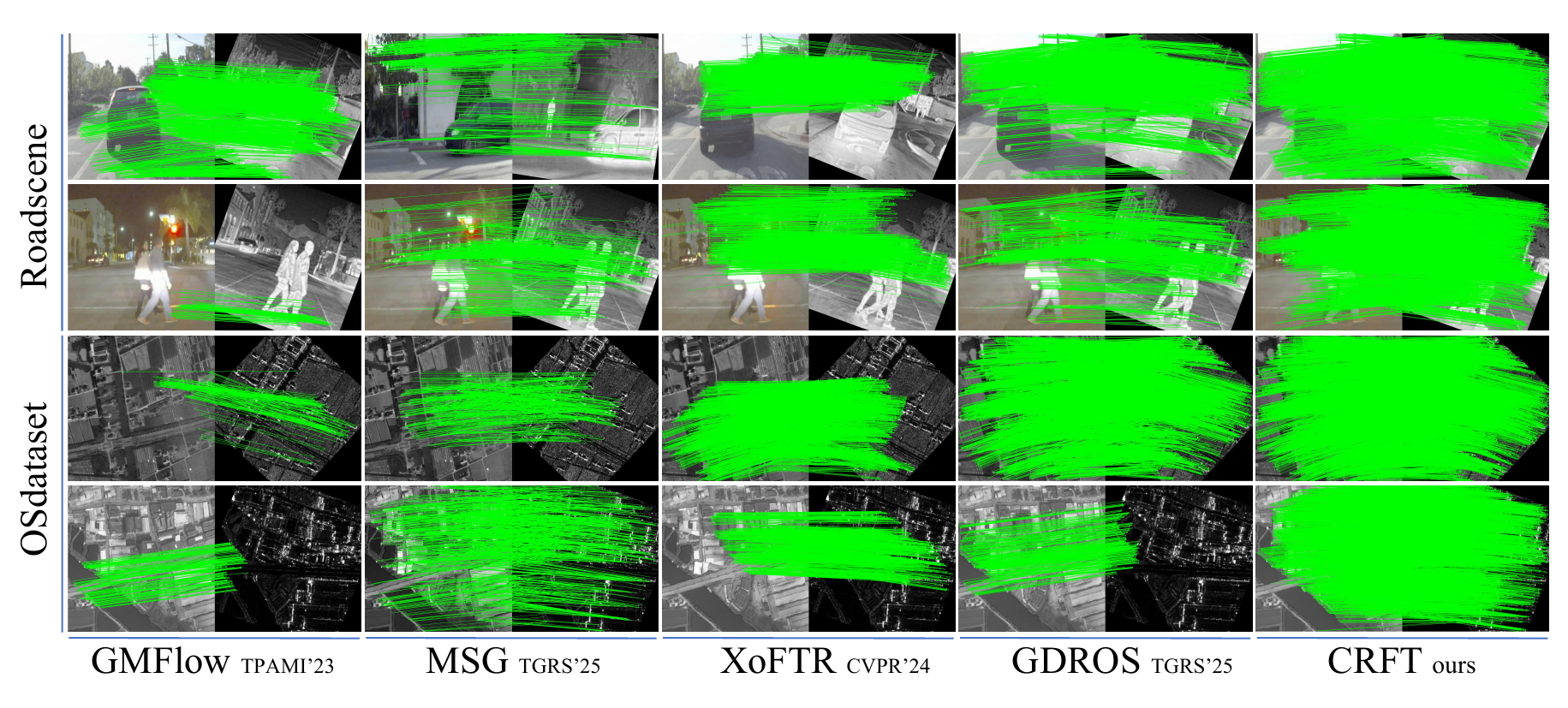} 
    \caption{\textbf{Quantitative comparison with state-of-the-art methods.} For each image pair, 5000 candidate correspondences are uniformly sampled, and only the matches with registration error below 2 pixels are visualized. The density and spatial consistency of the displayed inliers reflect the alignment accuracy of different methods across both OSdataset and RoadScene. CRFT produces the largest number of geometrically consistent inliers, demonstrating clear advantages in cross-modal and affine-deformed scenarios.
    }
    \label{fig:pointfig}
\end{figure*}

\begin{figure*}[t]
  \centering
  \begin{subfigure}{0.45\textwidth}
    \centering
    \includegraphics[width=\linewidth]{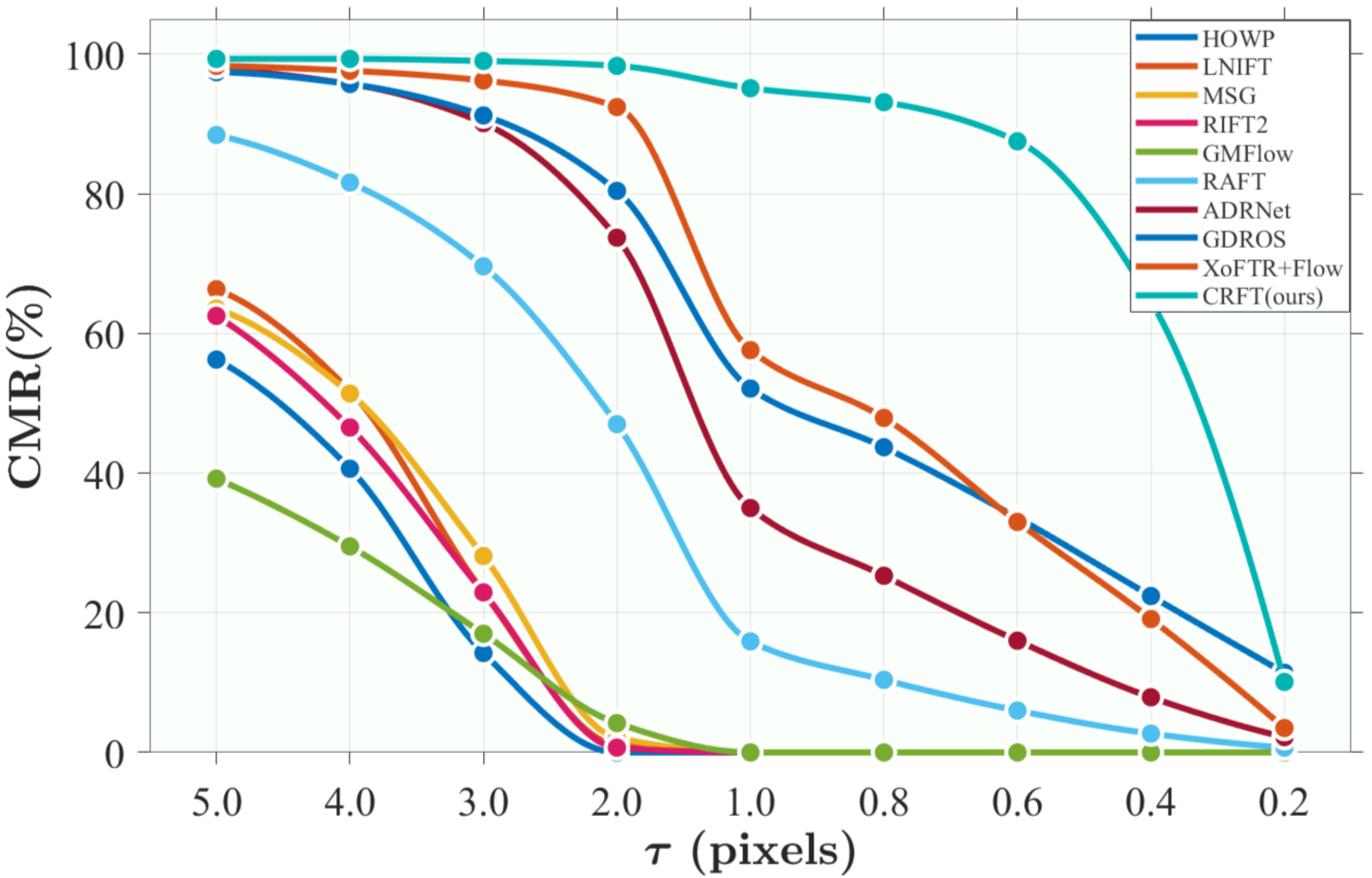}
    \caption{OSdataset}
    \label{fig:compareOS}
  \end{subfigure}
  \hfill
  \begin{subfigure}{0.45\textwidth}
    \centering
    \includegraphics[width=\linewidth]{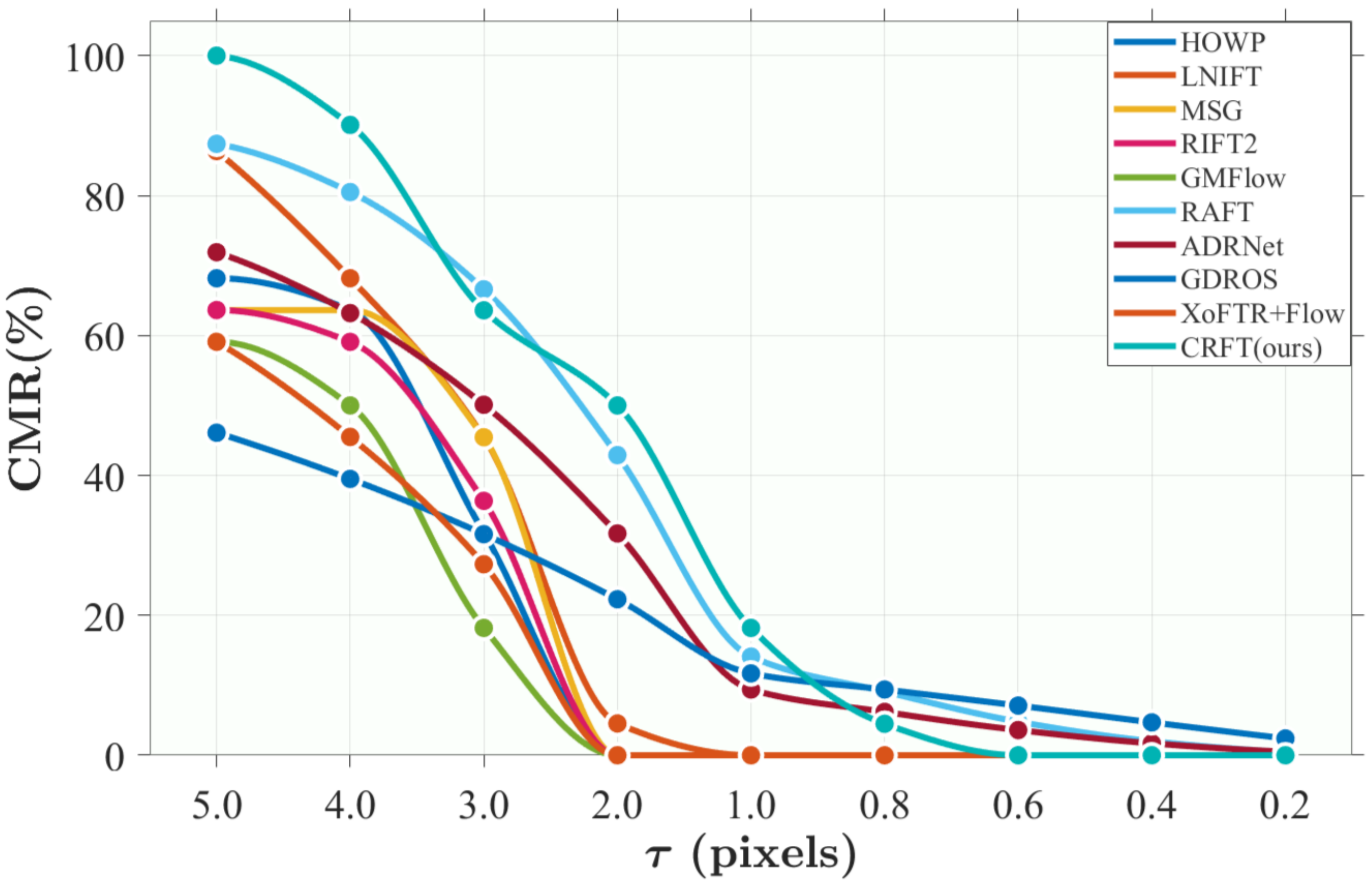}
    \caption{RoadScene}
    \label{fig:compareRoad}
  \end{subfigure}
  \caption{\textbf{CMR curves under varying thresholds on (a) OSdataset and (b) RoadScene.} We compare CRFT with representative handcrafted, sparse, and (semi-)dense matching methods by evaluating the CMR across a range of pixel thresholds. CRFT consistently achieves the highest CMR across a wide range of thresholds.}
  \label{fig:compare_both}
\end{figure*}

\begin{table}[t]
  \centering
  \footnotesize   
  \setlength{\tabcolsep}{3.5pt}
  \renewcommand{\arraystretch}{0.9}

  \caption{\textbf{Quantitative comparison on the OSdataset}. 
  AEPE and CMR at thresholds $t=\{3,1,0.7\}$. 
  Best results are in \textbf{bold}, and second-best are \underline{underlined}.}
  \label{tab:osdataset_results}

  \begin{tabular}{lcccc}
    \toprule
    \multirow{2}{*}{Methods} & \multirow{2}{*}{AEPE $\downarrow$} &
    \multicolumn{3}{c}{CMR $\uparrow$ (\%)} \\ 
    \cmidrule(lr){3-5}
     &  & $@3px$ & $@1px$ & $@0.7px$ \\
    \midrule
    HOWP~\cite{zhang2023histogram}~{\scriptsize ISPRS'23}  & 26.91 & 14.2 & 0.0 & 0.0 \\
    LNIFT~\cite{li2022lnift}~{\scriptsize TGRS'22}         & 48.18 & 22.9 & 0.0 & 0.0 \\
    MSG~\cite{zheng2025msg}~{\scriptsize TGRS'25}          & 44.38 & 28.1 & 0.0 & 0.0 \\
    RIFT2~\cite{li2019rift}~{\scriptsize TIP'20}           & 23.61 & 22.9 & 0.0 & 0.0 \\
    GMFlow~\cite{xu2023unifying}~{\scriptsize TPAMI'23}    & 11.91 & 17.0 & 0.0 & 0.0 \\
    RAFT~\cite{teed2020raft}~{\scriptsize ECCV'20}         & 3.51  & 69.6 & 15.9 & 8.7 \\
    ADRNet~\cite{xiao2024adrnet}~{\scriptsize TGRS'24}                          & 1.67  & 90.1 & 35.0 & 20.6 \\
    GDROS~\cite{sun2025gdros}~{\scriptsize TGRS'25}        & 1.34  & 91.1 & 49.2 & 35.5 \\
    XoFTR+Flow~\cite{tuzcuouglu2024xoftr}~{\scriptsize CVPR'24}  
    & \underline{1.13} & \underline{96.2} & \underline{57.6} & \underline{41.7} \\
    \midrule
    \textbf{CRFT (ours)} & \textbf{0.65} & \textbf{99.0} & \textbf{95.1} & \textbf{89.9} \\
    \bottomrule
  \end{tabular}

  \vspace{1mm}
\end{table}

\begin{table}[t]
  \centering
  \footnotesize   
  \setlength{\tabcolsep}{3.5pt}
  \renewcommand{\arraystretch}{0.9}
  \caption{\textbf{Quantitative comparison on the RoadScene}. 
  AEPE and CMR at thresholds $t=\{3,1,0.7\}$.}
  \label{tab:roadscene_results}

  \begin{tabular}{lcccc}
    \toprule
    \multirow{2}{*}{Methods} & \multirow{2}{*}{AEPE $\downarrow$} & 
    \multicolumn{3}{c}{CMR $\uparrow$ (\%)} \\ 
    \cmidrule(lr){3-5}
     &  & $@3px$ & $@1px$ & $@0.7px$  \\
    \midrule
    HOWP~\cite{zhang2023histogram}~{\scriptsize ISPRS'23}        
        & 22.01  & 31.8  & 0.0   & 0.0   \\
    LNIFT~\cite{li2022lnift}~{\scriptsize TGRS'22}         
        & 24.24  & 30.7  & 0.0   & 0.0   \\
    MSG~\cite{zheng2025msg}~{\scriptsize TGRS'25}           
        & 50.87  & 45.5  & 0.0   & 0.0   \\
    RIFT2~\cite{li2019rift}~{\scriptsize TIP'20}         
        & 17.27  & 36.4  & 0.0   & 0.0   \\
    GMFlow~\cite{xu2023unifying}~{\scriptsize TPAMI'23}        
        & 12.02  & 18.2  & 0.0   & 0.0   \\
    RAFT~\cite{teed2020raft}~{\scriptsize ECCV'20}          
        & 8.92   & \underline{66.6}  & \underline{14.1}  & \underline{8.0}   \\
    ADRNet~\cite{xiao2024adrnet}~{\scriptsize TGRS'24}     
        & \underline{4.72}   & 50.1  & 9.4   & 4.8   \\
    GDROS~\cite{sun2025gdros}~{\scriptsize TGRS'25}         
        & 8.81   & 31.6  & 11.7  & \textbf{8.8}   \\
    XoFTR+Flow~\cite{tuzcuouglu2024xoftr}~{\scriptsize CVPR'24}    
        & 4.83   & 27.3   & 0.0   & 0.0   \\
    \midrule
    \textbf{CRFT (ours)} 
        & \textbf{2.37}  & \textbf{68.2}  & \textbf{18.2}  & 4.5 \\
    \bottomrule
  \end{tabular}
  \vspace{1mm}
\end{table}

We evaluate CRFT on OSdataset and RoadScene, comparing it against ten representative handcrafted, sparse, semi-dense, and dense correspondence methods. To guarantee a strictly fair comparison, all evaluated baselines were fine-tuned prior to benchmarking. Across all settings, CRFT achieves the best AEPE, the highest CMR under strict thresholds, and the most stable performance under multimodal appearance shifts and geometric transformations.

\paragraph{Results on OSdataset.}
Table~\ref{tab:osdataset_results} and Fig.~\ref{fig:compareOS} show that CRFT achieves dominant performance on OSdataset. 
First, CRFT obtains an AEPE of 0.65, the only sub-pixel result among all evaluated methods, outperforming the second-best XoFTR+Flow (1.13) by 42.5\% and GDROS (1.34) by 51.5\%. 
More importantly, CRFT exhibits strong stability under strict thresholds: at $0.7$\,px it reaches a CMR of 89.9\%, which is 4.36$\times$ higher than ADRNet (20.6\%), 2.53$\times$ higher than GDROS (35.5\%), and 2.15$\times$ higher than XoFTR+Flow (41.7\%). 
Traditional feature-based methods show severe degradation under cross-modal shifts, with AEPE ranging from 17.27 to 50.87 and CMR falling to zero below 1\,px. Optical-flow baselines (GMFlow, RAFT) also fail under modality discrepancies, confirming that homogeneous-domain flow models struggle under large geometric deformation scenarios. 
Overall, CRFT is the only method maintaining high performance from relaxed to strict thresholds, demonstrating robust global-local matching ability on optical-SAR data.

\paragraph{Results on RoadScene.}
RoadScene introduces strong nonlinear intensity variations, blur, and viewpoint distortion across optical-infrared pairs, making it even more challenging. 
CRFT achieves the best AEPE of 2.37, outperforming ADRNet (4.72) by 49.8\% and GMFlow (12.02) by 80.3\%. 
In terms of CMR, CRFT achieves the highest value at the 3\,px threshold (68.2\%), slightly surpassing RAFT (66.6\%), and it leads by a clear margin at the stricter 1\,px threshold (18.2\% vs.\ 14.1\% for RAFT and 9.4\% for ADRNet). 
Under the $0.7$\,px threshold, most deep methods degrade sharply, while CRFT maintains competitive performance, reflecting the effectiveness of its discrepancy-guided refinement and SGT-based geometric reasoning. 
Handcrafted and sparse methods fail almost completely below 2\,px, consistent with their limited adaptability to large modality gaps.

\paragraph{Overall Discussion.}
Across both datasets, CRFT consistently surpasses the representative baselines in AEPE, CMR stability, and robustness to multimodal distortions. 
Its coarse-to-fine hierarchical matching stabilizes global alignment, while the proposed discrepancy-guided attention and SGT-based recurrent refinement enable precise correction of local geometric inconsistencies. 
These results demonstrate that CRFT provides a superior balance between global structural consistency and fine-grained local accuracy, setting a new performance level for multimodal image registration.

\subsection{Ablation Study}

\begin{figure}[t]   
    \centering  
    \centering    
    \includegraphics[width=0.9\linewidth]{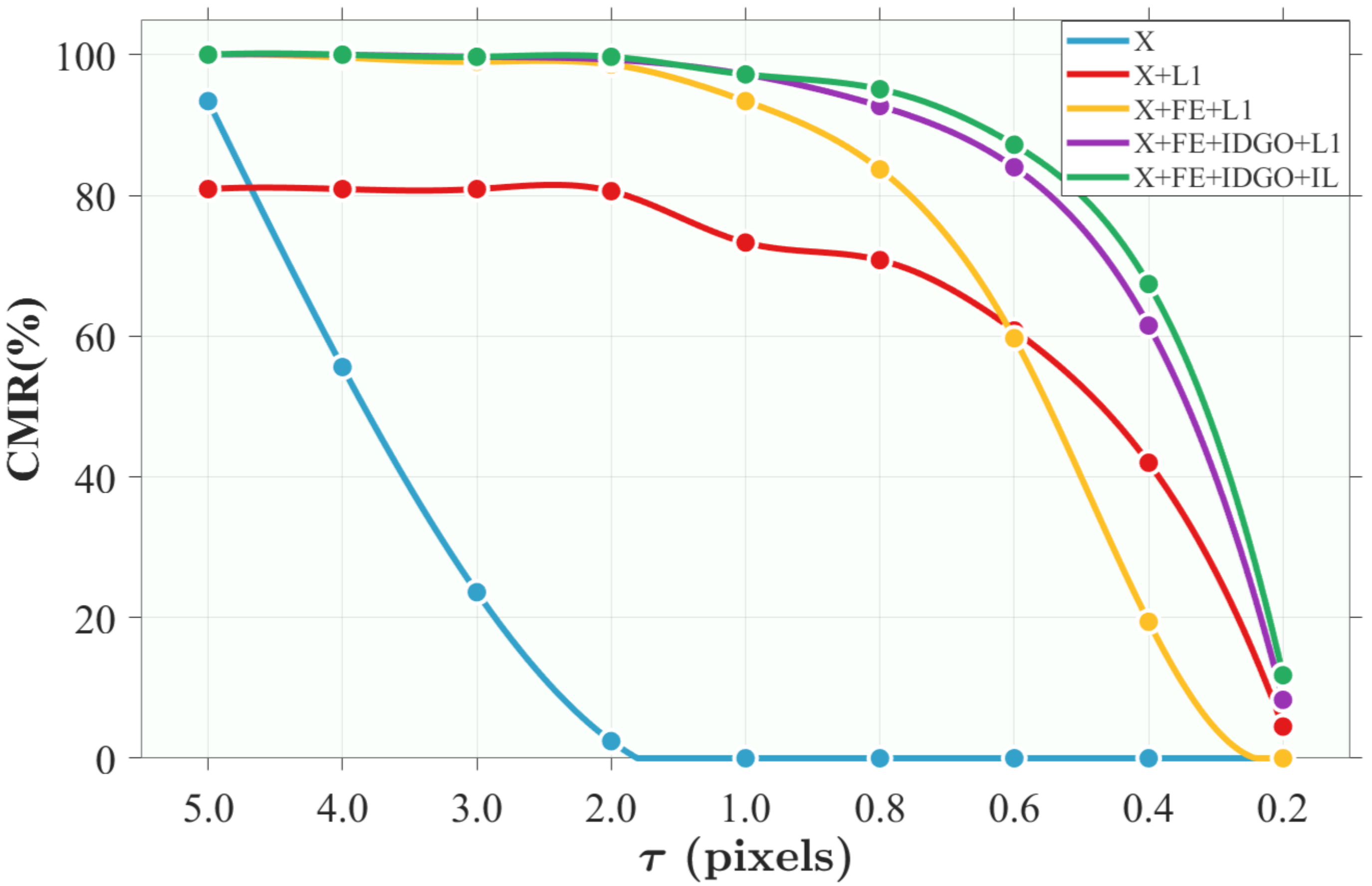} 
    \caption{\textbf{Ablation study on OSdataset.} We evaluate the contribution of each component in CRFT by progressively adding the flow estimation (FE), iterative discrepancy-guided optimization (IDGO), and the Iterative Loss (IL) to the XoFTR baseline. The CMR curves across varying thresholds demonstrate that each module brings a clear performance gain, while integrating all modules (X+FE+IDGO+IL) yields the most significant improvement, especially under strict pixel thresholds.}
    \label{fig:ablationCMR}
\end{figure}

We perform ablation experiments on the OSdataset using random translation (30 px), scaling within [0.9, 1.1], and rotation of $30^\circ$.  
A smaller rotation is used here (vs.\ $90^\circ$ in the main evaluation) because the XoFTR baseline fails under large rotations, making incremental analysis impossible.  
Starting from XoFTR (denoted as X), we progressively add the L1 loss, coarse Flow Estimation (FE), iterative discrepancy-guided optimization (IDGO), and the full iterative loss (IL).  
Results shown in Fig.~\ref{fig:ablationCMR} and Table~\ref{tab:ablation}.

\textbf{(1) Baseline (X).}  
The baseline performs poorly, with CMR falling to zero below 2 px, indicating its inability to capture reliable cross-modal fine-grained correspondences.

\textbf{(2) X+L1.}  
Adding dense L1 supervision significantly stabilizes training, reducing AEPE from 3.76 to 0.51 and improving CMR across all thresholds.

\textbf{(3) X+FE+L1.}  
Coarse-scale FE improves robustness at relaxed thresholds ($\tau > 1$), but performance still collapses under sub-pixel settings, showing that coarse correlation alone cannot achieve precise alignment.

\textbf{(4) X+FE+IDGO+L1.}  
Introducing IDGO yields substantial accuracy gains by iteratively refining local windows via SGT and discrepancy-guided attention.  
CMR@0.7px increases from 73.3\% to 88.9\%, confirming its effectiveness for fine-scale correction.

\textbf{(5) X+FE+IDGO+IL (CRFT).}  
The full CRFT achieves the best performance, obtaining the lowest AEPE (0.40) and highest sub-pixel accuracy (93.1\%@0.7px).  
The iterative loss further boosts the stability of late-stage refinements.

\noindent \textbf{Summary.} Each module contributes measurable improvements; IDGO is essential for sub-pixel precision, and the complete CRFT pipeline delivers the strongest accuracy and robustness.

\begin{table}[htbp]
    \centering
   \footnotesize   
   \setlength{\tabcolsep}{3.5pt}
   \renewcommand{\arraystretch}{0.9}

    \caption{\textbf{Ablation experiments on OSdataset}. 
    AEPE and CMR at thresholds $t=\{3,1,0.7\}$.}

    \begin{tabular}{lcccc}
    \toprule
    \multirow{2}{*}{Method} & \multirow{2}{*}{AEPE $\downarrow$} &
    \multicolumn{3}{c}{CMR $\uparrow$ (\%)} \\
    \cmidrule(l){3-5}
    & & $@3px$ & $@1px$ & $@0.7px$ \\ 
    \midrule
    (1) X & 3.76 & 23.6 & 0.0 & 0.0 \\
    (2) X+L1 & 0.51 & 80.9 & 73.3 & 66.7 \\
    (3) X+FE+L1 & 0.63 & \underline{99.0} & \underline{93.4} & 73.3 \\
    (4) X+FE+IDGO+L1 & \underline{0.43} & \textbf{99.7} & \textbf{97.2} & \underline{88.9} \\
    (5) X+FE+IDGO+IL (CRFT) & \textbf{0.40} & \textbf{99.7} & \textbf{97.2} & \textbf{93.1} \\
    \bottomrule
    \end{tabular}
    \label{tab:ablation}
\end{table}

\noindent\textbf{Efficiency Analysis.}
CRFT contains 11.96M parameters and is trained end-to-end on an RTX~4090 GPU in approximately 4.0 hours on OSdataset and 0.65 hours on RoadScene. CRFT is efficient with 11.96M parameters and 11.47 GFLOPs, running at 0.033s per pair. The DGFO module is lightweight; each iteration adds a marginal cost of 0.93M parameters, 1.90 GFLOPs, and 0.10 ms.

\section{Conclusion}
\label{sec:conclusion}

We presented CRFT, a consistent-recurrent feature flow Transformer that unifies coarse-to-fine matching and iterative geometric refinement for image registration.
CRFT learns modality-independent feature flow through multi-scale correlation and progressively refines alignment via discrepancy-guided optimization with Spatial Geometric Transform. This design enables robust correspondence estimation under severe appearance gaps, large rotations, scale changes, and affine distortions.
Extensive experiments on optical-SAR and optical-infrared datasets demonstrate that CRFT achieves state-of-the-art accuracy, particularly under strict sub-pixel thresholds, highlighting its strong geometric fidelity.
Beyond registration, CRFT offers a generalizable and scalable solution for multimodal spatial alignment with broad applicability across remote sensing, autonomous navigation, and medical imaging. Its modular design further ensures compatibility with future advances in large-scale vision models, geometric reasoning, and multimodal correspondence learning.

\section*{Acknowledgment}
\label{sec:Acknowledgment}

This work was supported by Hebei Natural Science Foundation: F2025501003, and the Fundamental Research Funds for the Central Universities: N2523014.

{
    \small
    \bibliographystyle{ieeenat_fullname}
    \bibliography{main}
}
\clearpage
\setcounter{page}{1}
\maketitlesupplementary
\appendix

\section{Visualization of Registration Results}

To provide deeper insight into the behavior of CRFT, we present additional qualitative visualizations on both OSdataset and RoadScene. These examples highlight the model’s robustness across heterogeneous modalities and challenging geometric conditions.

\begin{figure}[t]
    \centering
    \includegraphics[width=0.45\textwidth]{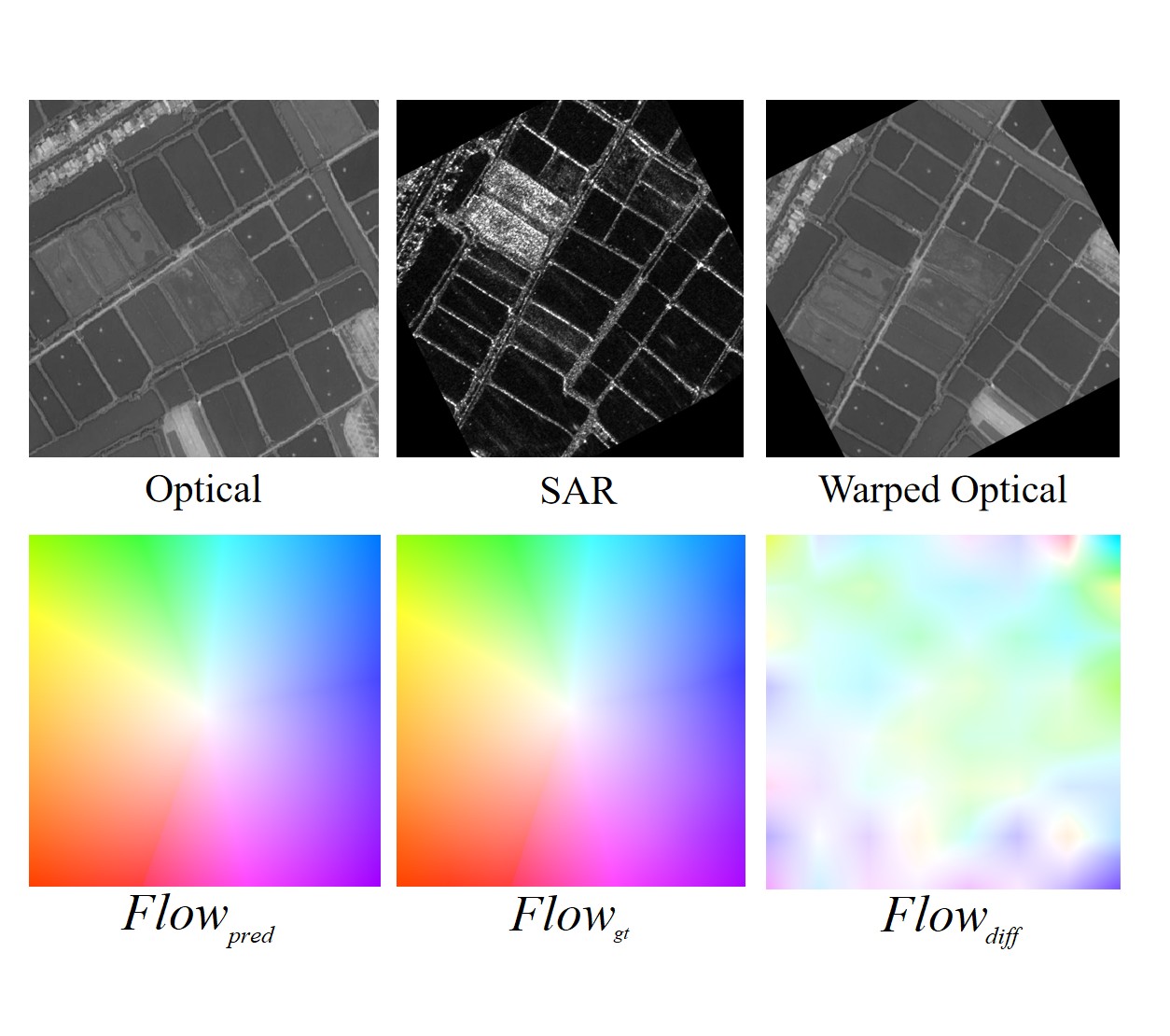}
    \caption{\textbf{Flow prediction visualization on OSdataset.} For each pair, we visualize the predicted dense flow, the ground-truth flow, and the corresponding difference map. 
    CRFT produces smooth and geometrically consistent flow fields across
    optical-SAR modality gap, indicating accurate geometric correspondence and robustness to strong appearance variations}
    \label{fig:OSflow}
\end{figure}

\begin{figure}[t]
    \centering
    \includegraphics[width=0.45\textwidth]{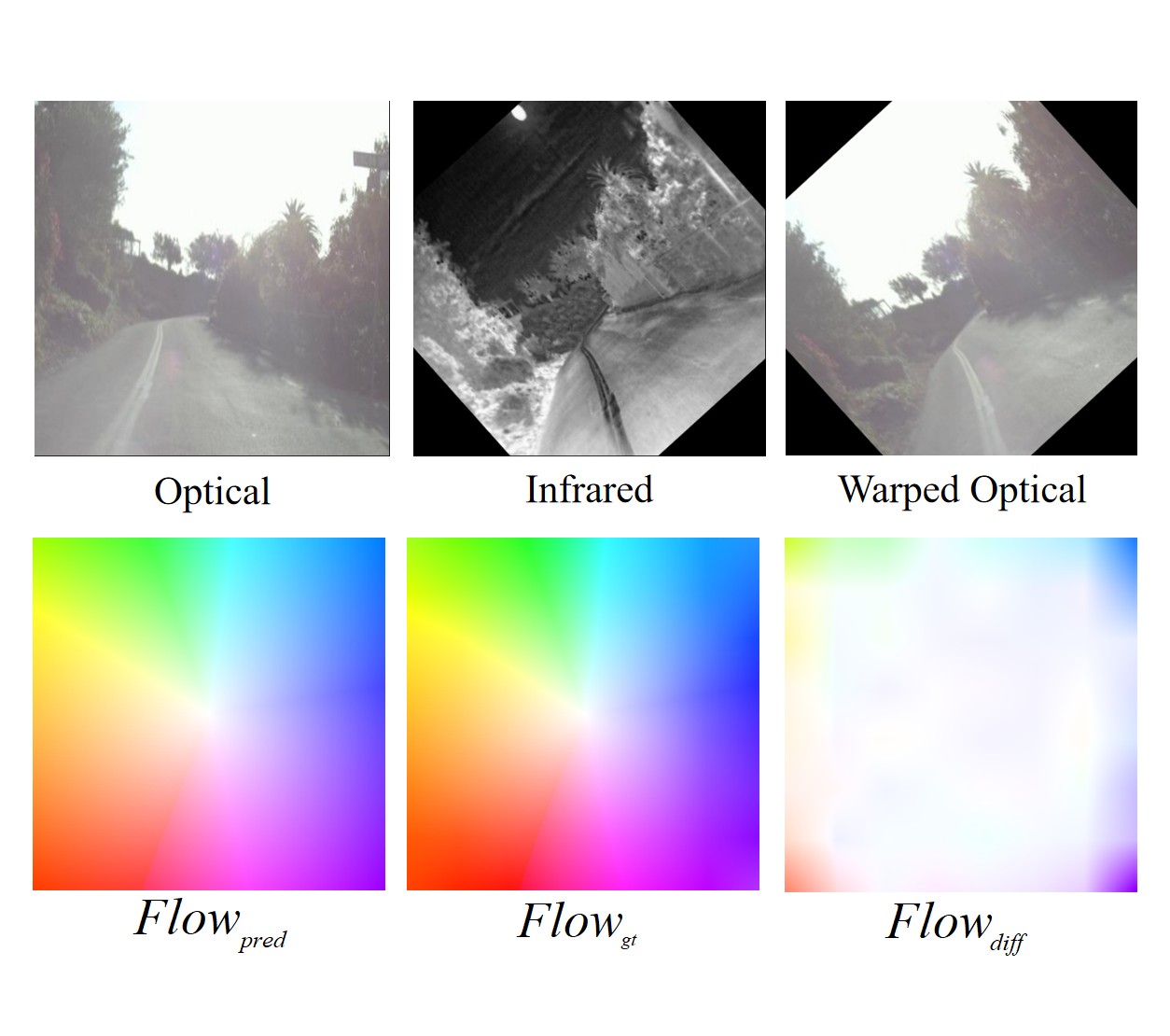}
    \caption{\textbf{Flow prediction visualization on RoadScene.} For each pair, we visualize the predicted dense flow, the ground-truth flow, and the corresponding difference map.     
    CRFT maintains stable and consistent flow behavior across heterogeneous modalities, demonstrating robustness to illumination changes, noise patterns, and texture inconsistencies in complex driving environments.}
    \label{fig:RSflow}
\end{figure}

\noindent
\textbf{Flow prediction visualization.}  
Figs.~\ref{fig:OSflow} and \ref{fig:RSflow} show dense flow predictions on OSdataset and RoadScene. CRFT produces smooth and structurally coherent flow fields that closely follow the scene geometry. The difference maps contain only small residuals, indicating accurate displacement estimation despite strong modality-induced appearance variations. These results confirm the stability of CRFT’s iterative discrepancy-guided flow refinement mechanism across different sensing modalities.

\begin{figure*}[t]
    \centering
    \includegraphics[width=0.85\textwidth]{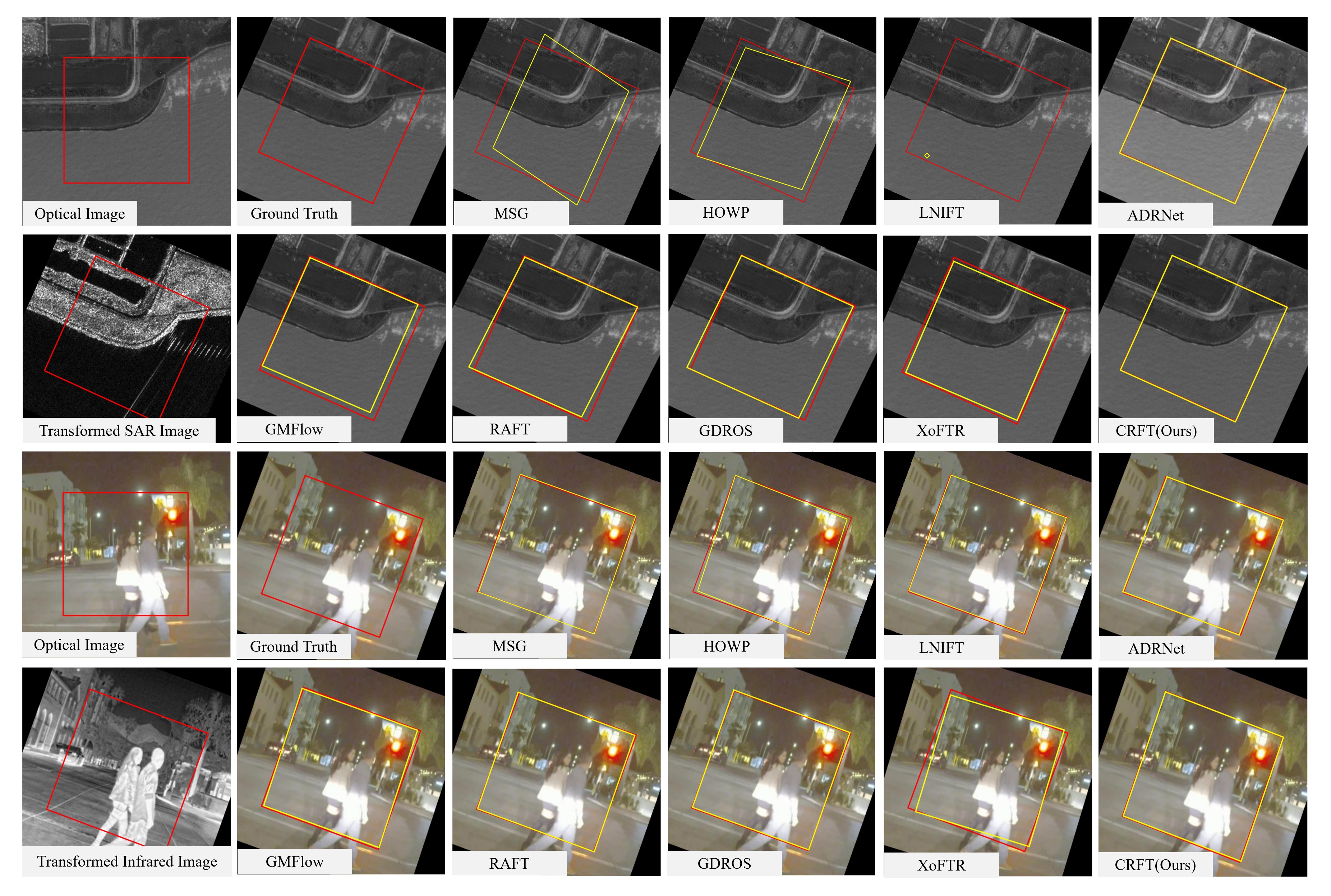}
    \caption{\textbf{Qualitative comparison on the OSdataset (top) and RoadScene (Bottom).} 
    The red quadrilateral denotes the ground-truth alignment, while the yellow quadrilateral shows the predicted registration result. CRFT achieves more accurate and geometrically consistent registration under large geometric deformation and modality gaps compared with existing optical-SAR registration baselines.}
    \label{fig:Rectangle}
\end{figure*}

\noindent
\textbf{Registration comparison across modalities.}  
Fig.~\ref{fig:Rectangle} presents large-scale qualitative comparisons on both OSdataset (top) and RoadScene (bottom). 
Classical methods such as HOWP and MSG, as well as deep learning-based approaches including GMFlow and XoFTR, exhibit reduced robustness under large rotations, scale variations, and substantial optical-SAR/infrared appearance gaps, often resulting in angular or translational misalignments. In contrast, CRFT produces registration results that align closely with the ground-truth quadrilaterals, yielding lower end-point errors and more stable geometric fidelity.

\begin{figure*}[t]
    \centering
    \includegraphics[width=0.8\textwidth]{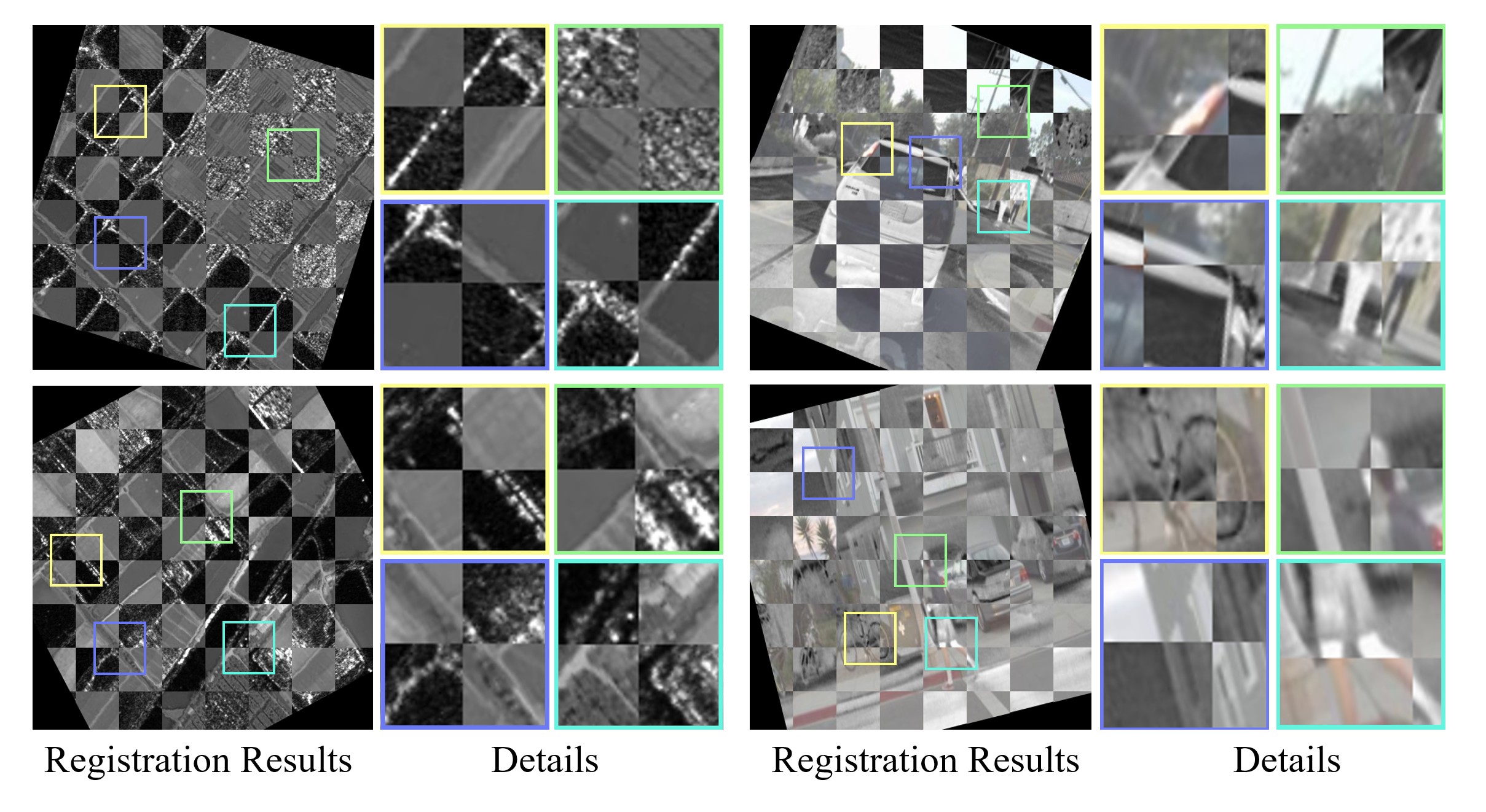}
    \caption{\textbf{Checkerboard registration on OSdataset (left) and RoadScene (right).} CRFT yields geometrically coherent checkerboard fusion across both optical-SAR and optical-infrared modalities. Zoomed-in regions demonstrate precise alignment of boundaries and textures, confirming the model’s ability to recover fine-scale geometric correspondence under large cross-modal appearance differences.}
    \label{fig:checkerboard}
\end{figure*}

\noindent
\textbf{Checkerboard fusion and fine-scale alignment.}  
To further evaluate pixel-level correspondence, Fig.~\ref{fig:checkerboard} visualizes checkerboard fusion results on OSdataset (left) and RoadScene (right). CRFT achieves coherent grid alignment across both datasets without tearing or ghosting artifacts. The zoomed-in patches demonstrate precise matching of fine structural elements such as field boundaries, edges, and textures. These observations validate CRFT’s ability to recover fine-scale geometric correspondence in complex cross-modal registration scenarios.

\section{Limitations and Failure Cases. }While our method demonstrates strong robustness under standard augmentations, we acknowledge a performance drop under extreme geometric distortions. when subjected to extreme $90^\circ$ rotations coupled with random scaling in the range of $[0.5,1.5]$, the AEPE on the OSdataset rises significantly to $9.75$. This reveals a vulnerability to extreme affine variations, highlighting the need for scale- and rotation-invariant representations in future research.

\end{document}